\documentclass[aps,prl,twocolumn,groupedaddress,showpacs,floatfix]{revtex4-1}

\usepackage{dcolumn}
\usepackage{amsmath}
\usepackage{amssymb}
\usepackage{graphicx}
\usepackage{bm}
\usepackage[T1]{fontenc}
\usepackage{color}
\usepackage{url}
\usepackage[bf]{subfigure}
\usepackage{rotating}
\usepackage{soul}
\usepackage{scalefnt}
\usepackage{algorithm2e}
\usepackage{hyperref}

\usepackage{color}
\definecolor{redcolor}{rgb}{1.0,0.,0.}

\definecolor{bluecolor}{rgb}{0.,0.,1.0}

\begin{document}

\title{Topic segmentation via community detection in complex networks}

\author{Henrique F. de Arruda}
\affiliation{Institute of Mathematics and Computer Sciences, University of S\~ao Paulo, S\~ao Carlos, S\~ao Paulo, Brazil}
\author{Luciano da F. Costa}
\affiliation{S\~ao Carlos Institute of Physics,\\ University of S\~ao Paulo, S\~ao Carlos, S\~ao Paulo, Brazil}
\author{Diego R. Amancio}
\email{diego@icmc.usp.br}
\affiliation{Institute of Mathematics and Computer Sciences, University of S\~ao Paulo, S\~ao Carlos, S\~ao Paulo, Brazil }


\begin{abstract}
Many real systems have been modelled in terms of network concepts, and written texts
are a particular example of information networks. In recent years, the use of network methods to analyze
language has allowed the discovery of several interesting findings, including the proposition of novel models
to explain the emergence of fundamental universal patterns.
While syntactical networks, one of the most prevalent networked models of written texts, display both scale-free and small-world
properties, such representation fails in capturing other textual features, such as the organization in topics or subjects.
In this context, we propose a novel network representation whose main purpose is to capture the semantical
relationships of words in a simple way. To do so, we link all words co-occurring in the same semantic context, which is defined
in a threefold way.
We show that the
proposed representations favours the emergence of communities of semantically related words, and this feature may be used to
identify relevant topics. The proposed methodology to detect topics was applied to segment selected Wikipedia articles. We have found that, in general, our methods outperform traditional bag-of-words representations, which
suggests that a high-level textual representation may be useful to study semantical features of texts.
\end{abstract}

\maketitle

\setcounter{secnumdepth}{1}

\section{Introduction}
\label{introduction}

With the ever-increasing amount of information available online, the classification of texts has established itself as one of the most relevant applications
supporting the development of efficient and informative search engines~\cite{Manning:1999}. As the correct categorization of texts is of chief importance for the success of any search engine, this task has been extensively studied, with significant success in some
scenarios~\cite{cla,herrera}. However, the comprehension of texts in a human-like fashion remains a challenge, as it is
the case of semantical analyses performed in disambiguating systems and sentiment analysis~\cite{Golder30092011,Navigli:2009}. While humans are able to identify
the content and semantics conveyed in written texts in a artlessly manner, automatic classification methods fail in several aspects mainly because the lack of a
general representation of world knowledge. Currently, the automatic categorization of texts stands as one of the most studied applications in information
sciences. In recent years, multidisciplinary concepts have been applied to assist the classification, including those based on physical
approaches~\cite{herrera,PhysRevE.92.022813,Amancio2012427,cansyntax,Delanoe201493}.

An important subarea of the research performed in text classification is the detection of subtopics in documents~\cite{Reynar:1999:SMT,Lancichinetti2015}. This
task is not only useful to analyze the set of subjects approached in the document, but also to assist computational applications, such as automatic
text summarization~\cite{refinp} and text recommendation~\cite{Adomavicius:2005}. Several approaches have been proposed to tackle the subtopic identification
problem. The most common feature employed to cluster together related sentences/paragraphs is the frequency of shared words~\cite{cooley,Blei:2003}. While most
of the works on this area considers only the presence/absence of specific words in sentences/paragraphs, only a few studies have considered the organization of
words to classify documents~\cite{stylometry}. Specifically, in this paper, we approach the text segmentation problem using the organization of words as a
relevant feature for the task. The non-trivial relationships between concepts are modelled via novel networked representations.

Complex networks have been used to analyze a myriad of real systems~\cite{survapp}, including several features of the human language~\cite{Cong2014598}. A
well-known model for text representation is the so-called word adjacency (co-occurrence) model, where two words are linked if they appear as neighbors in the
text. Though seemingly simple, this model is able to capture authors' styles~\cite{AmancioAltmann,euJst15,Mehri20122429}, textual
complexity~\cite{cansyntax,amancomp} and many other textual aspects~\cite{Cong2014598,Masuci1,Masuci2}. A similar model, referred to as syntactical network, takes into account the syntactical representation of texts by connecting words syntactically related. Syntactical networks shared topological properties of other real world networks, including both scale-free and small-word behaviors. These networks have been useful
for example to capture language-dependent features~\cite{PhysRevE.69.051915}. Unlike traditional models, in this paper, we extend traditional text
representations to capture semantical features so that words are connected if they are related semantically.  Here, we take the view that a given subtopic is
characterized by a set of words which are internally connected, with a few external relationships with words belonging to other subjects. As we shall show, this
view can be straightforwardly translated into the concept of community structure, where each semantical structure can be regarded as a different network
community~\cite{Fortunato201075}. The proposed framework was evaluated in a set of Wikipedia articles, which are tagged according to their subjects.
Interestingly, the network approach turned to be more accurate than traditional method in several studied datasets, suggesting that the
structure of subjects can be used to improve the task. Taken together, our proposed methods could be useful to analyse written texts in a multilevel networked
fashion, as revealed by the topological properties obtained from traditional word adjacency models and networks of networks in higher hierarchies via community
analysis.

This manuscripts is organized as follows. In Section \ref{sec:CN}, we present the proposed network representation to tackle the subtopic identification task.
In Section \ref{sec:database} we present the dataset employed for the task. The results obtained are then shown in Section \ref{sec:res}.
Finally, in Section \ref{sec:conc}, we conclude this paper by presenting perspectives for further research on related areas.

\section{Complex network approach}
\label{sec:CN}

The approach we propose here for segmenting texts according to subjects relies upon a networked representation of texts. In this section, we also detail the
method employed for identifying network communities, for the proposed methodology requires a prior segmentation of text networks.


\subsection{Representing texts as networks} \label{sec:net}

A well-known representation of texts as networks is the word-adjacency model, where each different word is a node and links are established between adjacent
words~\cite{Cancho2261,amancioMachine}. This model, which can be seen as a simplification of the so-called syntactical networks~\cite{PhysRevE.69.051915}, has been used with success to identify styles in applications related to authorship attribution, language identification and authenticity verification~\cite{sciento,voynich1,voynich2}. The
application of this model in clustering analysis is not widespread because traditional word adjacency networks are not organized in communities, if one
considers large pieces of texts.  In this paper, we advocate that there is a strong relationship between communities structures and subjects. For this reason,
we modify the traditional word adjacency model to obtain an improved representation of words interactions in written texts. More specifically, here we present a
threefold extension which considers different strategies of linking non-adjacent words.

Prior to the creation of the model itself, some pre-processing are applied. First, we remove \emph{stopwords}, i.e. the very frequent words conveying no
semantic meaning are removed (e.g. \emph{to}, \emph{have} and \emph{is}). Note that, such words may play a pivotal role in style identification, however in this
study they are not relevant because such words are subject independent. The pre-processing step is also responsible for removing other non-relevant tokens, such
as punctuation marks. Because we are interesting in representing concepts as network nodes, it is natural to consider that variant forms of the same word
becomes a unique node of the network. To do so, we lemmatize the words so that nouns and verbs are mapped into their singular and infinitive
forms~\cite{miller1995wordnet}. To assist this process, we label each word according to their respective parts of speech in order to solve possible ambiguities.
This part of speech labeling is done with the high-accurate maximum entropy model defined in~\cite{ratnaparkhi1996maximum, malecha2010maximum}. Finally, we
consider only nouns and verbs, which are the main classes of words conveying semantical meanings {\cite{hindle1990noun,bechet2014combine}}.
%
%
In all proposed models, each remaining word (lemmatized nouns and verbs) becomes a node. Three variations concerning the establishment of links between
such words are proposed:


\begin{itemize}
\item  \emph{Extended co-occurence (EC) model:} in this model, the edges are established between two words if they are separated by $d=(\omega-1)$ or less intermediary words in the pre-processed text.  For example, if $\omega=2$ and the pre-processed
text comprises the sentence ``$w_1$ $w_2$ $w_3$ $w_4$'', then the following set of edges is created: $\mathcal{E} = \{w_1 \leftrightarrow w_2, w_2
\leftrightarrow w_3, w_3 \leftrightarrow w_4, w_1 \leftrightarrow w_3, w_2 \leftrightarrow w_4\}$. Note that, if $w=1$, the traditional co-occcurrence (word
adjacency) model is recovered, because only adjacent words are connected.
%

\item \emph{Paragraph-based (PB) model:}
in this model, we link in a clique the words belonging to the same paragraph. We disregard, however, edges linking words separated by more than $d$
intermediary words in the pre-processing text.
This method relies on the premise that paragraphs represents the fundamental sub-structure in which a text is organized, whose words are semantically
related~\cite{Veronis2004223}.

\item  \emph{Adapted paragraph-based (APB) model:} the PB model does not take into account the fact that words may co-occur in the same paragraph just by
chance, as it is the case of very frequent words. To consider only significant links in the PB model, we test the statistical significance of co-occurrences
with regard to random, shuffled texts~\cite{martinez2011disentangling}. Given two words $v_i$ and $v_j$, an edge is created only if the frequency of
co-occurrences ($k$) (i.e. the number of paragraphs in which $v_i$ and $v_j$ co-occurs) is much higher than the same value expected in a null model. The significance of the frequency $k$ can be computed using the quantity $p(k)$,
which quantifies the probability of two words to appear in the same paragraph considering the null model. To compute the probability $p(k)$, let $n_1$ and $n_2$
be the number of distinct partitions in which $v_i$ and $v_j$ occur, respectively. The distribution of probability for $k$, the number of co-occurrences between $v_i$ and $v_j$,  can be
computed as
\begin{equation}  \label{feq} \nonumber
        p(k)= \frac{(N;k,n_1-k,n_2-k)}{(N;n_1)(N;n_2)},~\textrm{where}
\end{equation}
\begin{equation}
	(x;y_1,\ldots,y_n) \equiv \frac{x!}{x_1! \ldots x_n!} \frac{1}{(x-y_1-\ldots y_n)!}. \nonumber
\end{equation}
The above expression for $p(k)$ can be rewritten in a more convenient way, using the notation
\begin{equation}
\{a\}_b \equiv \prod_{i=0}^{b-1} (a-i),
\end{equation}
which is adopted for $a \geq b$. In this case, the likelihood $p(k)$ can be written as
\begin{widetext}
\begin{eqnarray} \label{opv}
\centering
    p(k) & = & \frac{\{n_1\}_k \{n_2\}_k \{N-n_1\}_{n_2-k}}{\{N\}_{n_2}\{k\}_k}
    = \frac{\{n_1\}_k \{n_2\}_k  \{N-n_1\}_{n_2-k}}{\{N\}_{n_2-k}\{N-n_2+k\}_k\{k\}_k } \nonumber \\
    & = & \prod_{j=0}^{n_2-k-1} \Bigg{[} \frac{N - j- n_1}{N-j} \Bigg{]} \prod_{j=0}^{k-1} \frac{(n_1-j)(n_2-j)}{(N-n_2+k-j)(k-j)}.
\end{eqnarray}
\end{widetext}
If in a given text the number of co-occurrences of two words is $r$, the $p$-value $p$ associated to $r$ can be computed as
%
\begin{equation} \label{eq.pvalue}
   p({k \geq r}) = \sum_{k \geq r} p(r),
\end{equation}
where $p(r)$ is computed according to equation \ref{opv}. Now, using equation \ref{eq.pvalue}, the most significant edges can
be selected.
%
 \end{itemize}

\begin{figure*}
\center
\includegraphics[width=1\textwidth]{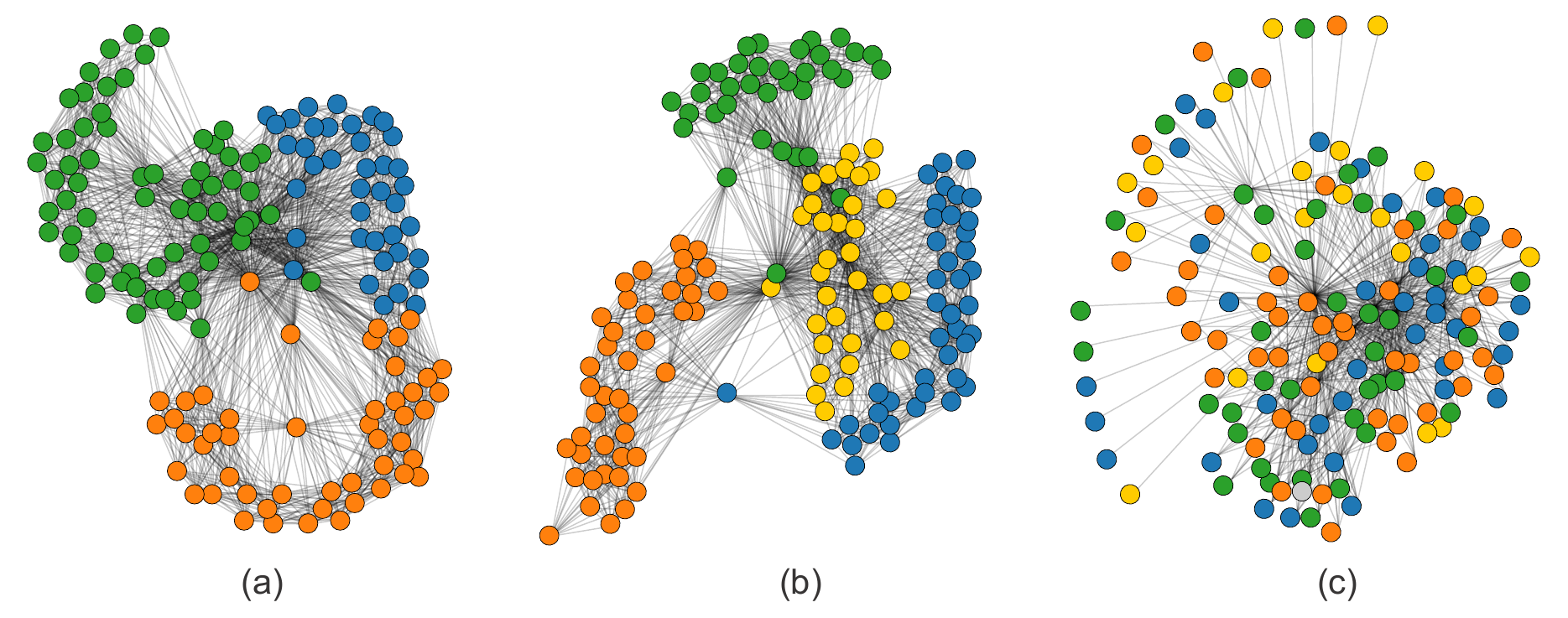}
\caption{Example of networks obtained from the following models: (a) EC, (b) PB and (c) APB. The colours represent in this case the community to which each node belongs. To obtain the partitions, we used the fast greedy algorithm \cite{clauset2004finding}. Note that the organization of nodes in communities varies from model to model. While in (a) and (b) the organization in communities is clear, in (c) the modular organization is much more weak.}
\label{fig:net1}
\end{figure*}

Figure~\ref{fig:net1} illustrates the topology obtained for the three proposed models representing a text with four paragraphs from Wikipedia~\footnote{\url{en.wikipedia.org/wiki/Car}}. Note that the
structure of communities depends on the method chosen to create the netswork. Especially, a lower modularity has been found for the APB model in this example and for this reason the organization in communities is not so clear. As we shall show, the identification of communities of extreme importance for identifying accurately the subjects.

%

\subsection{From network communities to text subjects}
\label{sec:communities}

The first step for clustering texts according to subjects concerns the computation of network communities, i.e. a region of nodes with several intra-edges (i.e.
edges inside the community) and a few inter-edges  (i.e. edges leaving the community). Methods for finding network communities have been applied in several
applications~\cite{arenas2004community,guimera2005worldwide,palla2005uncovering}, including in text analysis~\cite{Londhe:2014}. The quality of partitions
obtained from community structure detection methods can be obtained from the modularity $Q$, which is defined as
\begin{equation} \label{eq.modularity}
	Q = \frac{1}{2M} \sum_{i} \sum_j \left( a_{ij} - \frac{k_i k_j}{2M}\right) \delta(c_i,c_j),
\end{equation}
where $a_{ij}$ denotes an element of the adjacent matrix (i.e. $a_{ij}=1$ if $i$ and $j$ are linked and $a_{ij}=0$, otherwise),  $c_i$ represents the membership of
the $i$-th node, $k_i = \sum_j a_{ij}$ is the node degree, $M = 1/2 \sum_i \sum_j a_{ij}$ is the total number of edges in the network and $\delta(x,y)$ is the
Kronecker's delta. Several algorithms use the modularity to assist the identification of partitions in networks. Note that the modularity defined in equation \ref{eq.modularity} quantifies the difference between the actual number of intra-links (i.e. the links in the same community, $a_{ij} \delta(c_i,c_j)$) and the expected number of intra-links (i.e.
the number of links in the same community of a random network, $( k_i k_j / 2M) \delta(c_i,c_j)$).
Here we use a simple yet efficient approach devised in~\cite{clauset2004finding}, where the authors define a greedy optimization strategy for $Q$. More
specifically, to devise a greedy optimization the authors define the quantities
\begin{equation}
	e_{ij} = \frac{1}{2M} \sum_v \sum_w a_{vw} \delta( c_v, i) \delta(c_w, j),
\end{equation}
which represents the fraction of edges that link nodes in community $i$ and $j$, and
\begin{equation}
	\alpha_i =  \frac{1}{2M} \sum_v k_v \delta(c_v,i),
\end{equation}
which denotes the fraction of ends of edges that are linked to nodes in community $i$. The authors redefine $Q$ in equation \ref{eq.modularity} by replacing
$\delta(c_v,c_w)$ to $\sum_i \delta(c_v,i) \delta(c_w,i)$:
\begin{eqnarray} \label{eq.mod2}
	Q & = & \frac{1}{2M} \sum_v \sum_w \Bigg{[} a_{vw} - \frac{k_v k_w}{2M} \Bigg{]} \sum_i \delta(c_v,i) \delta(c_w,i) \nonumber \\
	    & = & \sum_i \Bigg{[} \frac{1}{2M}  \sum_v \sum_w a_{vw} \delta(c_v,i) \delta(c_w,i) \nonumber \\
	    & & - \frac{1}{2M} \sum_v k_v \delta(c_v,i) \frac{1}{2M} \sum_w k_w \delta (c_w,i) \Bigg{]} \nonumber \\
	    & = & \sum_i (e_{ii} - \alpha_i^2).
\end{eqnarray}
Using the modularity obtained in equation \ref{eq.mod2}, it is possible to detect communities using an agglomerative approach. First, each node is associated to a distinct
community. In other words, the partition initially comprises only singleton clusters. Then, nodes are joined to optimize the variation of modularity, $\Delta
Q_{ij}$. Thus, after the first agglomeration, a multi-graph is defined so that communities are represented as a single node in a new adjacency matrix with
elements $a_{ij}' = 2M e_{ij}$. More specifically, the operation of joining nodes is optimized by noting that $\Delta Q_{ij} > 0$ only if communities $i$ and
$j$ are adjacent. The implementation details are provided in~\cite{clauset2004finding}.
Note that our approach does not rely on any specific community identification method. We have specifically chosen the fast-greedy method because, in
preliminary experiments, it outperformed other similar algorithms, such as the multi-level approach devised in~\cite{blondel2008fast} (result not shown).

%

\begin{figure*}
\center
\includegraphics[width=1\textwidth]{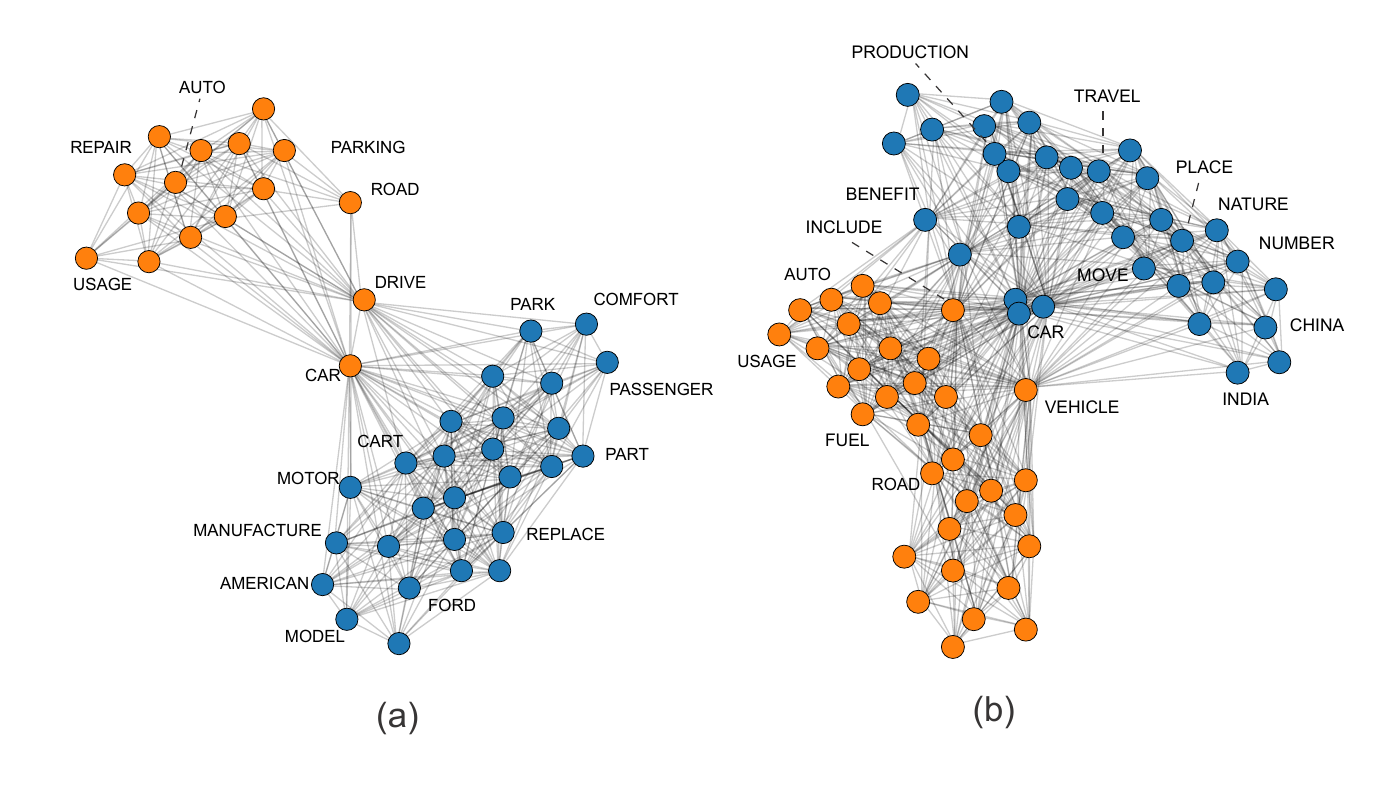}
\caption{Example of community obtained with the following models: (a) EC; and (b) PB. As expected, most of the words belong to  a unique subject, while a few
words lies at the overlapping region. The visualization of communities obtained with the APB model is not as good as the ones obtained with both EC and PB
methods.}
\label{fig:net2}
\end{figure*}

Given a partition of the network established by the community detection method, we devised the following approach for clustering the text in $n_s$ distinct
subjects. Let $\mathcal{C} = \{c_1,c_2,\ldots\}$ and $\Pi=\{\pi_1,\pi_2,\ldots\}$ be the set of communities in the network and the set of paragraphs in the text, respectively.
If the obtained number of communities ($n_c$) is equal to the expect number of subjects $n_s$, then we assign the label $c_i$ to the word that
corresponds to node $i$ in the text. As a consequence, each paragraph is represented by a set of labels $\mathcal{L}(\pi_j) = \{ l_1^{(\pi_j)},l_2^{(\pi_j)},\ldots\}$, where $l_i^{(\pi_j)} \in \mathcal{C}$ is the label associated to the $i$-th word of the $j$-th paragraph ($\pi_j$). The number of occurrences of each label in each paragraph is defined as
\begin{equation}
	f(c_i,\pi_j) = \sum_{l \in \mathcal{L}(\pi_j)} \delta(c_i,l).
\end{equation}
Thus the subject associated to the paragraph $\pi_j$ is
\begin{equation} \label{oassunto}
	\tilde{s}(\pi_j) = \arg \max_{c_i \in \mathcal{C}} f(c_i,\pi_j),
\end{equation}
i.e. the most frequent label is chosen to represent the subject of each paragraph. To illustrate the application of our method, we show in Figure \ref{fig:net2}
the communities obtained in a text about cars, whose first paragraph approaches the definition and invention of cars, and the remaining paragraphs
present their parts.

The expression in equation \ref{oassunto} is suitable to be applied only when $n_c = n_s$. Whenever the number of expected subjects is lower than the number of
network communities, we adopt a strategy to reduce the quantity of communities found. To do so, the following algorithm is applied:

\begin{algorithm}[H]
 \KwData{$n_c$, the number of communities and $n_s$, the number of subjects.}
 \KwResult{A match between communities and subjects is established.}
 \While{ $(n_c > n_s)$}{
  $c_k$ = label of the community with the largest {overlapping region}\;
  erase all nodes from $c_k$\;
  detect again the community structure of the network formed of the remaining nodes\;
  update $n_c$\;
 }
\end{algorithm}

According to the above algorithm, if the most overlapping community is to be estimated, a measure
to quantify the degree of overlapping must be defined. In the present work, we defined an overlapping
index, $\sigma$, which is computed for each community. To compute $\sigma(c_i)$, we first recover
all paragraphs whose most frequent label is the one associated to community $c_i$. These paragraphs
are represented by the subset $\Pi_D(c_i) =\{ \pi \in \Pi~|~\tilde{s}(\pi) = c_i \}$.  Next, we count how many
words in $\Pi_D(c_i)$ are associated with the community $c_i$.  The overlapping is then defined as the
the amount of words in $\Pi_D(c_i)$ which are associated to a community $c_j \neq c_i$. Mathematically,
the overlapping is defined as
\begin{equation}
	\sigma(c_i)= 1 - \sum_{\substack{\pi \in \\ \Pi_D(c_i)}} \sum_{\substack{l \in \\\mathcal{L}(\pi)}} \delta(c_i,l).
\end{equation}

To evaluate the performance of the methods, we employed the following methodology. Let $s(\pi_i)$ be the subject associated to the $i$-th paragraph according
to Wikipedia. Here, we represent each different subtopic as a integer number, so that $s(\pi_i) \in [1,n_s]$.
Let $\tilde{s}(\pi_i) \in [1,n_s]$ be the label associated to the $i$-th paragraph according to a specific clustering method, as defined in equation
\ref{oassunto}. To quantify the similarity of two sets $S = \{s(\pi_1),s(\pi_2),\ldots\}$ and $\tilde{S} = \{\tilde{s}(\pi_1),\tilde{s}(\pi_2),\ldots\}$, it is
necessary consider all combinations of labels permutations either on  $S$ or  $\tilde{S}$, because the same partition may be defined with distinct labels. For
example, if $n_s = 2$,  $S=\{1,1,2,2\}$ and $\tilde{S}=\{2,2,1,1\}$, both partitions are the same, even if a straightforward comparison (element by element)
yields a zero similarity.   To account for distinct labelings in the evaluation, we define the operator $\mathcal{P}$, which maps a sequence of labels to all
possible combinations. Thus, if $S = \{1,1,2,2\}$, then
\begin{equation}
	\mathcal{P}(S) = \{ \{1,1,2,2 \}, \{2,2,1,1 \} \}. \nonumber
\end{equation}
Equivalently, the application of the $\mathcal{P}$ to ${S}$ yields two components:
\begin{equation}
	\mathcal{P}({S},1)=\{1,1,2,2\} \textrm{ and }
	\mathcal{P}({S},2)=(2,2,1,1). \nonumber
\end{equation}
The accuracy rate, $\Gamma$, is then defined as
\begin{equation}
	\Gamma = \max_i \mathcal{H}( S, \mathcal{P}(\tilde{S},i) ),
\end{equation}
where $\mathcal{H}(X,Y)$ is the operator that compares the similarity between two subsets
$X = \{x_1,x_2,\ldots\}$ and $Y = \{y_1,y_2,\ldots\}$ and is mathematically defined as:
\begin{equation}
	\mathcal{H}(X,Y) = \sum_i \delta(x_i,y_i).
\end{equation}


%
%

\section{Database}
\label{sec:database}

The artificial dataset we created to evaluate the methods is formed from paragraphs extracted from Wikipedia articles.
The selected articles can be classified in $8$ distinct topics and 5 distinct subtopics:
\begin{enumerate}

\item {\bf Actors}: Jack Nicholson, Johnny Depp, Robert De Niro, Robert Downey Jr. and Tom Hanks.

\item {\bf Cities}: Barcelona (Spain), Budapest (Hungary), London (United Kingdom), Prague (Czech Republic) and Rome (Italy).

\item {\bf Soccer players}:  Diego Maradona (Argentina), Leonel Messi (Argentina), Neymar Jr. (Brazil), Pel\'e (Brazil) and Robben (Netherlands).

\item {\bf Animals}: bird, cat, dog, lion and snake.

\item {\bf Food}: bean, cake, ice cream, pasta, and rice.

\item {\bf Music}:  classical, funk, jazz, rock and pop.

\item {\bf Scientists}: Albert Einstein, Gottfried Leibniz, Linus Pauling, Santos Dumont and Alan Turing.

\item {\bf Sports}: football, basketball, golf, soccer and swimming.

\end{enumerate}
To construct the artificial
texts, we randomly selected paragraphs from the articles using two parameters: $n_s$, the total number of subtopics addressed in the artificial text; and
$n_p$, the number of paragraphs per subtopic. For each pair $(n_s, n_p)$ we have compiled a total of $200$ documents. To create a dataset with distinct
granularity of subtopics, the random selection of subtopics was performed in a two-fold way. In the dataset comprising coarse-grained subtopics, hereafter
referred to as CGS dataset, each artificial text comprises $n_s$ subtopics of \emph{distinct} topics. Differently, in the dataset encompassing fine-grained
subtopics, hereafter referred to as FGS dataset, the texts are made up of distinct subtopics of the \emph{same} major topic.

\section{Results}
\label{sec:res}

In this section, we analyze the statistical properties of the proposed models in terms of their organization in communities. We also evaluate the performance of
our model in the artificial dataset and compare with more simple models that do not rely on any networked information.
%

\subsection{Modularity analysis} \label{sec.mod}

The models we propose to cluster topics in texts relies on the ability of a network to organize itself in a modular way. For this reason, we study how the modularity
of networks depends on the models parameters. The unique parameter that may affect {networks modularity in our models is $\omega$, which amounts to the distance of links in texts (see definition in the description of the EC model in Section \ref{sec:CN}). Note that the distance $\omega$ controls the total number of edges of the network, so a comparison of networks modularity for distinct values of $\omega$ is equivalent to comparing networks with distinct average degrees. Because the average degree plays an important role on the computation of the modularity~\cite{Fortunato201075}, we defined here a normalized modularity $Q_N$ that only takes into account the organization of the
network in modules and is not biased towards dense networks. The normalized modularity $Q_N$ of a given network with modularity $Q$ is computed as
\begin{equation}
	Q_N = Q - \langle Q_S \rangle,
\end{equation}
where $Q_S$ is the average modularity obtained in $30$ equivalent random networks with the same number of nodes and edges of the original
network.

In Figure \ref{fig:mod}, we show the values of $Q$, $Q_N$ and $Q_S$ in the dataset {FGS}. In the EC model, high values of modularity $Q$ are found for
low-values of $\omega$. To the best of our knowledge, this is the first time that a modular structure is found in a traditional word adjacency network (i.e. the
EC model with $\omega=1$) in a relatively small network.
As $\omega$ takes higher values and the network becomes more dense, the modularity $Q$ decreases. In a similar way, the average modularity $\langle Q_S
\rangle$ obtained in equivalent random networks also decreases when $\omega$ increases. A different behavior arises for the normalized modularity $Q_N$, as
shown in Figure~\ref{fig:mod}(a). The modularity $\langle Q_S \rangle$ initially takes a low value, reaching its maximum when $\omega=\omega_{max}=20$. Note
that when $\omega>\omega_{max}$, there is no significant gain in modularity $Q_N$. A similar behavior appears in the other two considered models (PB and APB) as
the normalized modularity becomes almost constant for $\omega>20$. Because in all models the normalized modularity takes high values for $\omega=20$, 
we have chosen this value to construct the networks for the
purpose of clustering topics. Even though a higher value of normalized modularity was found for $\omega>20$ in Figures~\ref{fig:mod}(b)-(c), we have decided not
to use larger values of $\omega$ in {these models because only a minor improvement in the quality is obtained for $\omega>20$}.


\begin{figure*}[]
\center
\subfigure[ref1][]{\includegraphics[width=5.9cm]{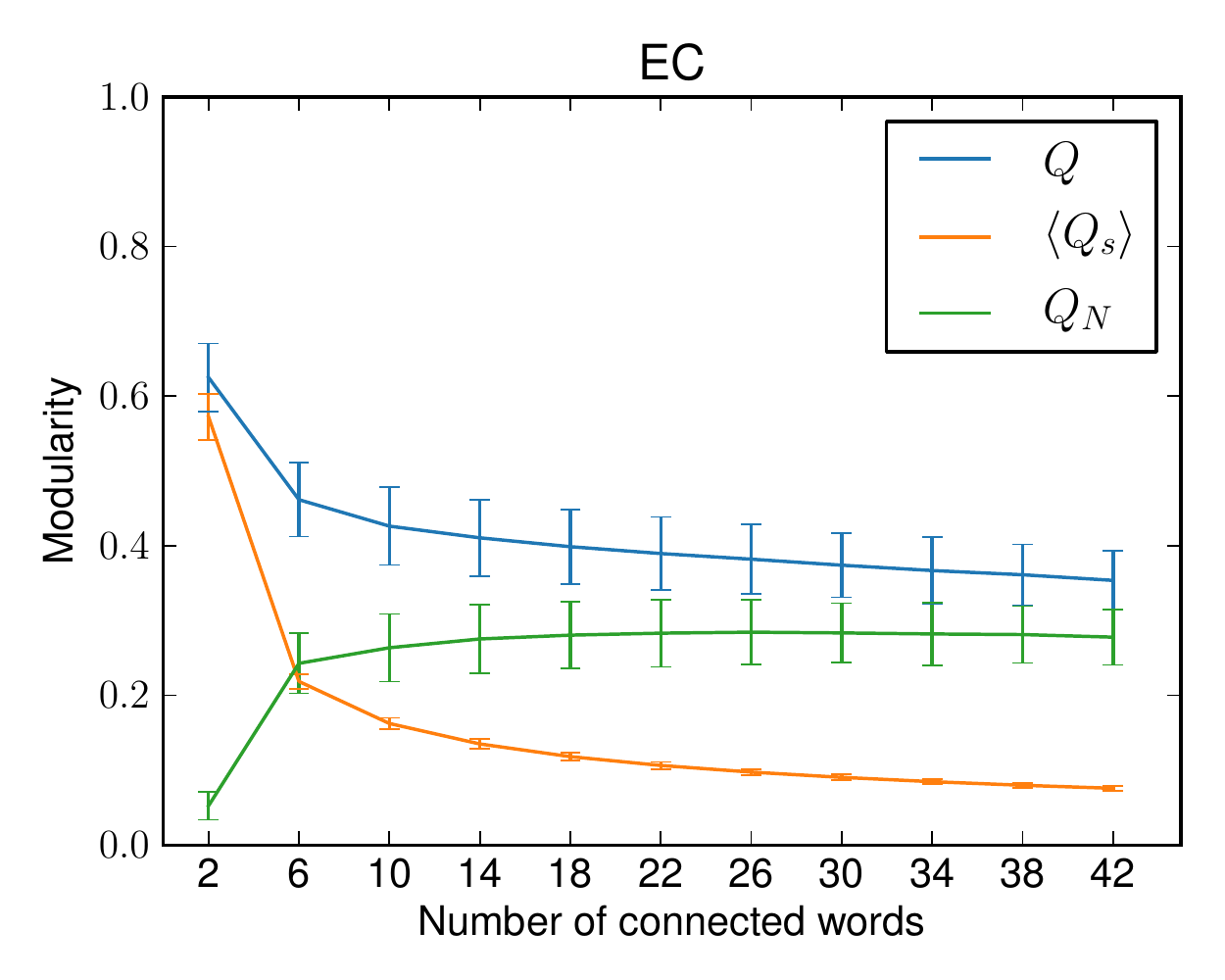}}
\subfigure[ref2][]{\includegraphics[width=5.9cm]{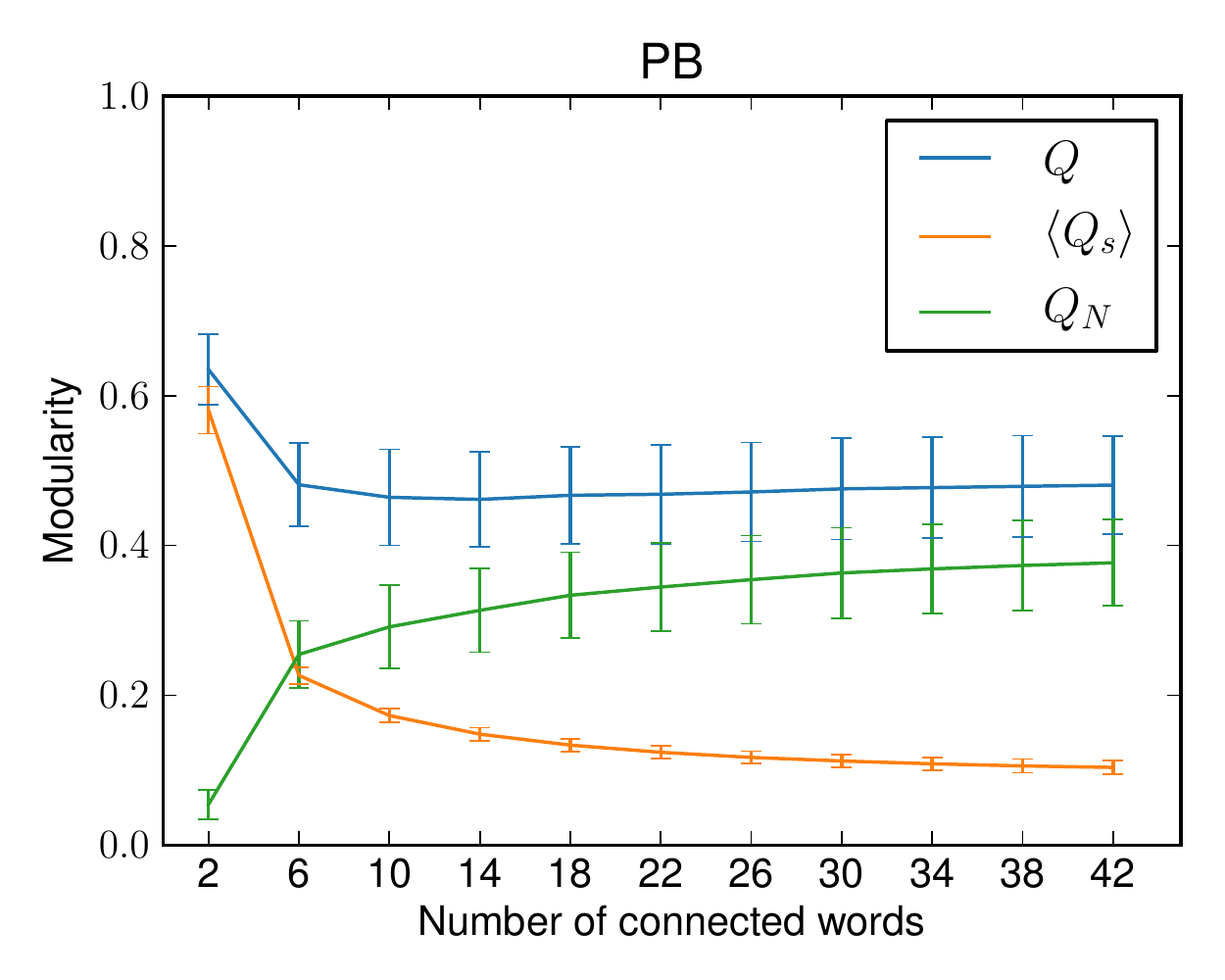}}
\subfigure[ref2][]{\includegraphics[width=5.9cm]{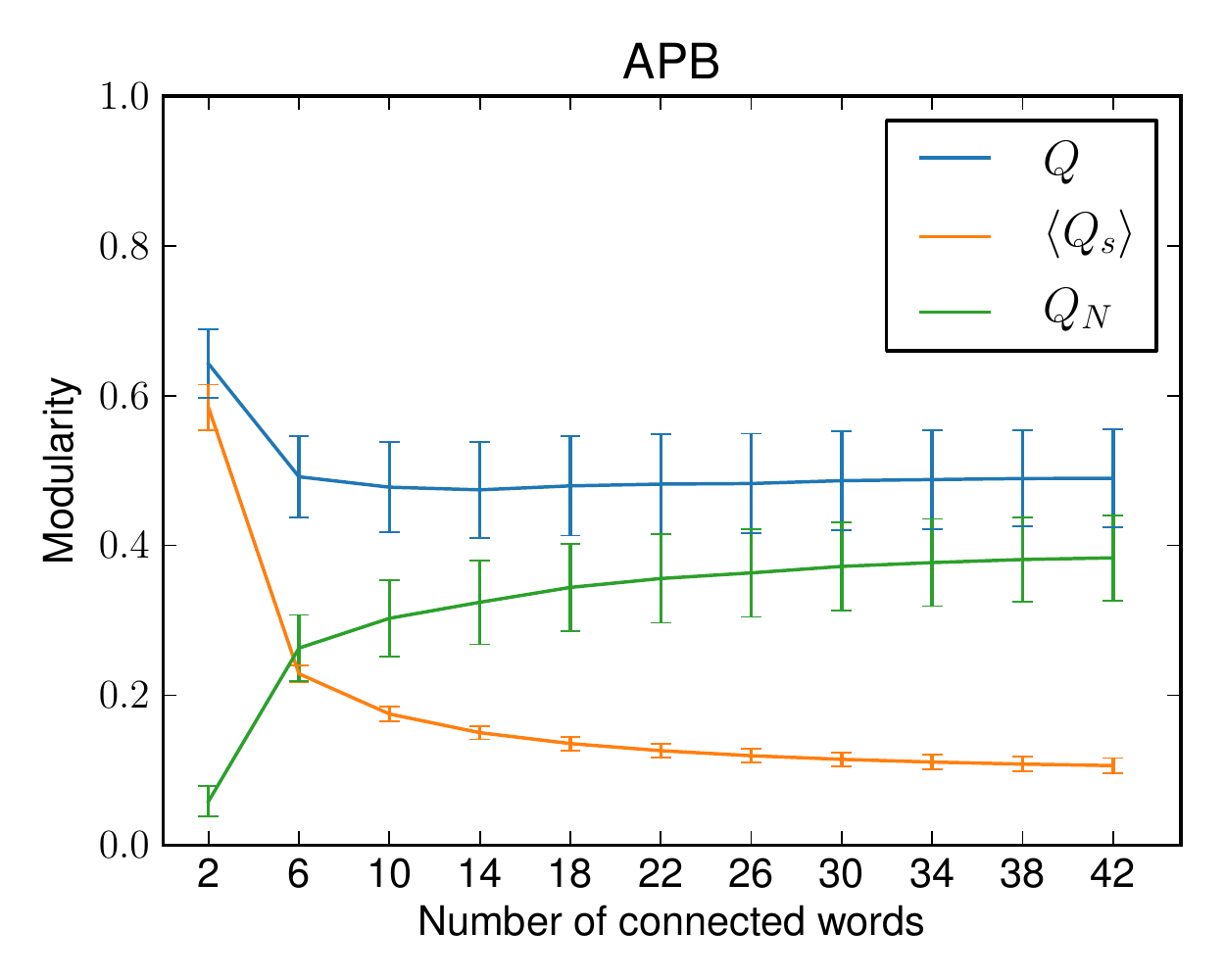}}
\caption{Example of modularity evolution using the proposed text representations (EC, PB and APB). The results were obtained in both CGS and FGS
datasets (see Section \ref{sec:database}). Note that the normalized modularity $Q_N$  does not display a significant increase for $\omega > 20$.}
\label{fig:mod}
\end{figure*}

\subsection{Performance analysis} \label{sec.perf}

To evaluate the performance of the proposed network-based methods, we evaluate the accuracy of the generated partitions in the {CGS and FDS} datasets presented in Section \ref{sec:database}. We have created two additional methods based on linguistic attributes to compare the performance of networked and non-networked methods. Both methods are based on a \emph{bag-of-words} strategy~\cite{Manning:1999}, where each paragraph is represented by the frequency of appearance of its words. To cluster the paragraphs in distinct subjects, we used two traditional clustering methods: k-means~\cite{Kanungo:2002:EKC:628329.628801} and Expectation
Maximization~\cite{Chai2015454}. In our results, the bag-of-words strategy based on the k-means and Expectation Mazimization algorithms are referred to as
BOW-K and BOW-EM, respectively.
%

In Figure \ref{fig:mod2}, we show the results obtained in the dataset comprising texts whose subtopics belong to distinct major topics ({CGS} dataset). We
classified the dataset in terms of the number of distinct subtopics in each text ($n_s$) and the total number of paragraphs per subtopic ($n_p$) (see Section
\ref{sec:database}). We first note that, in all cases, at least one networked approach outperformed both  BOW-K and BOW-EM approaches. In some cases, the
improvement in performance is much clear (see Figure \ref{fig:mod2} (d), (e) and (f)), while in others the difference in performance is lower. When comparing
all three networked approaches, the APB strategy, in average, performs better than others when the number of subtopics is $n_s \leq 3$. For $n_s=4$, both EC and
PB strategies outperforms  the APB method.
\begin{figure*}[]
\center
\subfigure[ref1][$n_s=2$ and $n_p=3$]{\includegraphics[width=5.9cm]{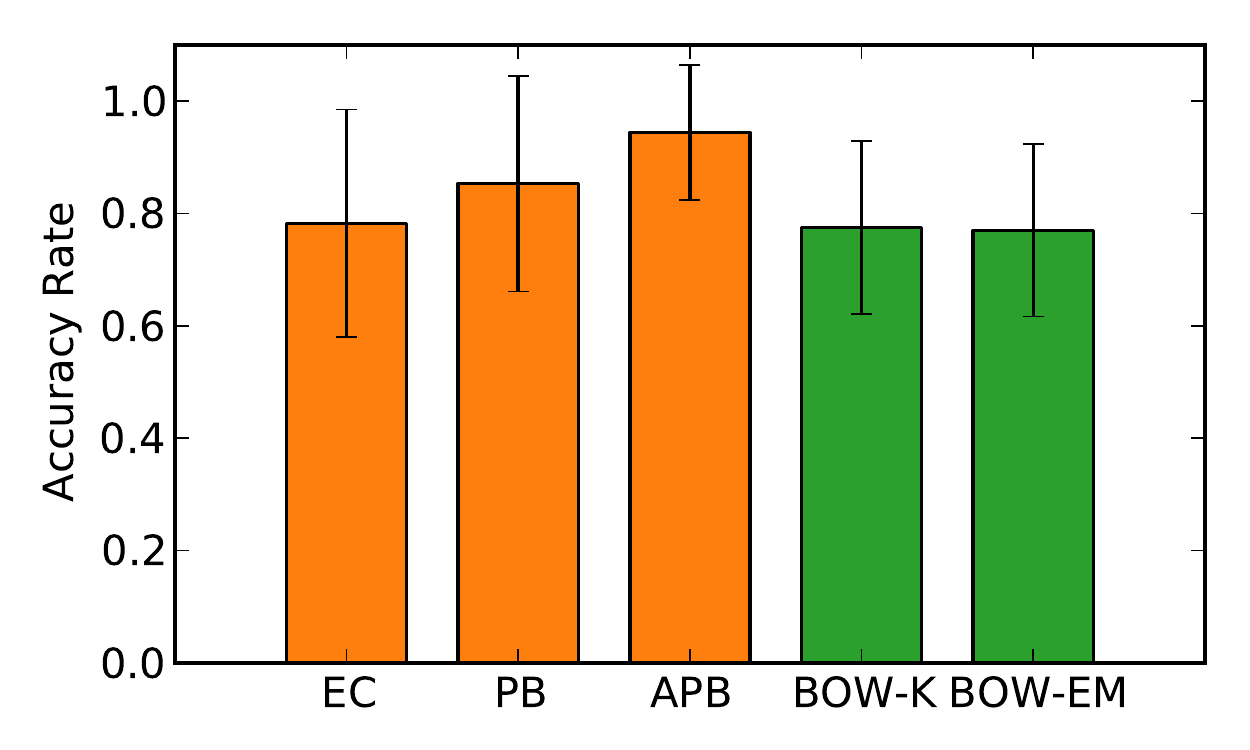}}
\subfigure[ref2][$n_s=2$ and $n_p=4$]{\includegraphics[width=5.9cm]{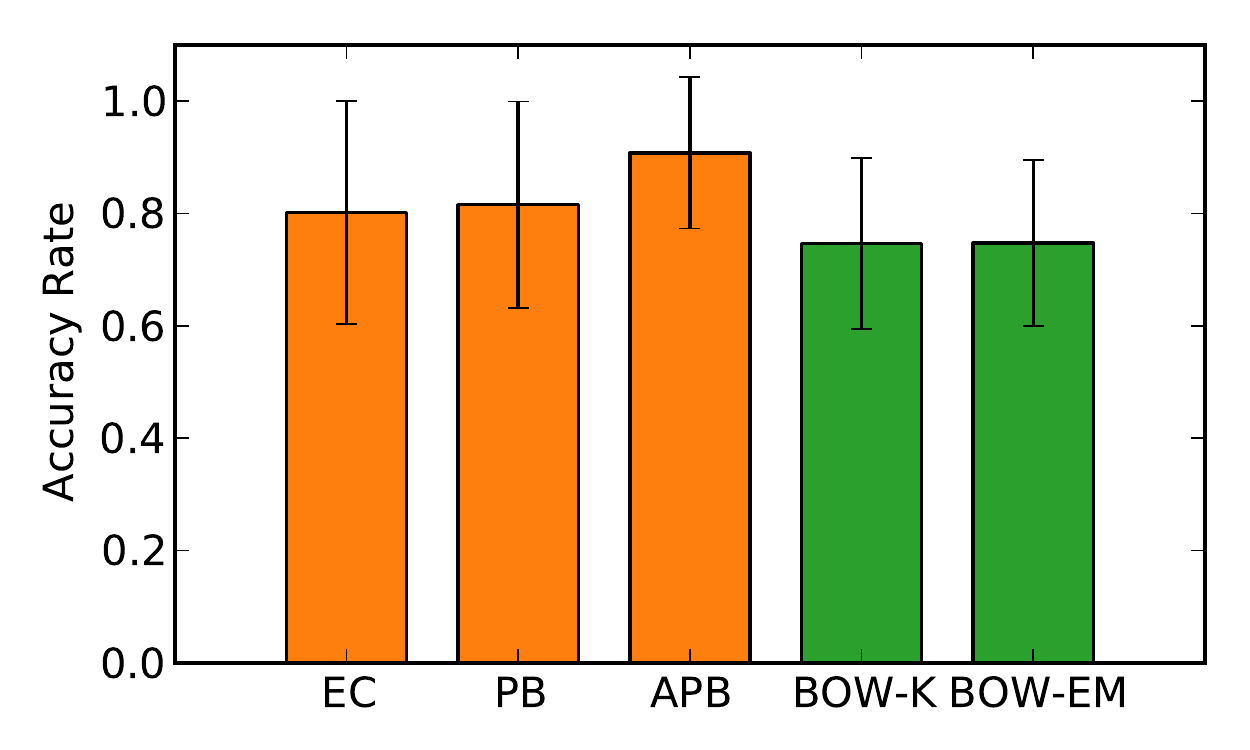}}
\subfigure[ref2][$n_s=2$ and $n_p=5$]{\includegraphics[width=5.9cm]{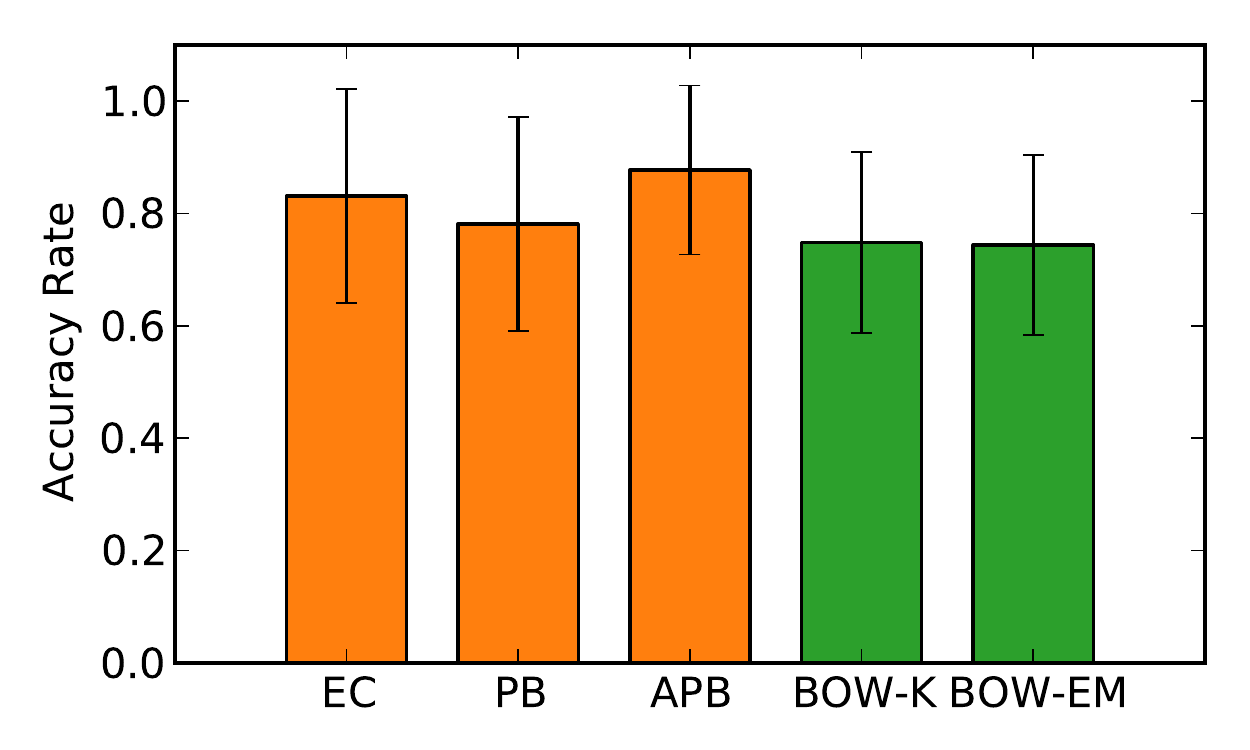}}
\subfigure[ref1][$n_s=3$ and $n_p=3$]{\includegraphics[width=5.9cm]{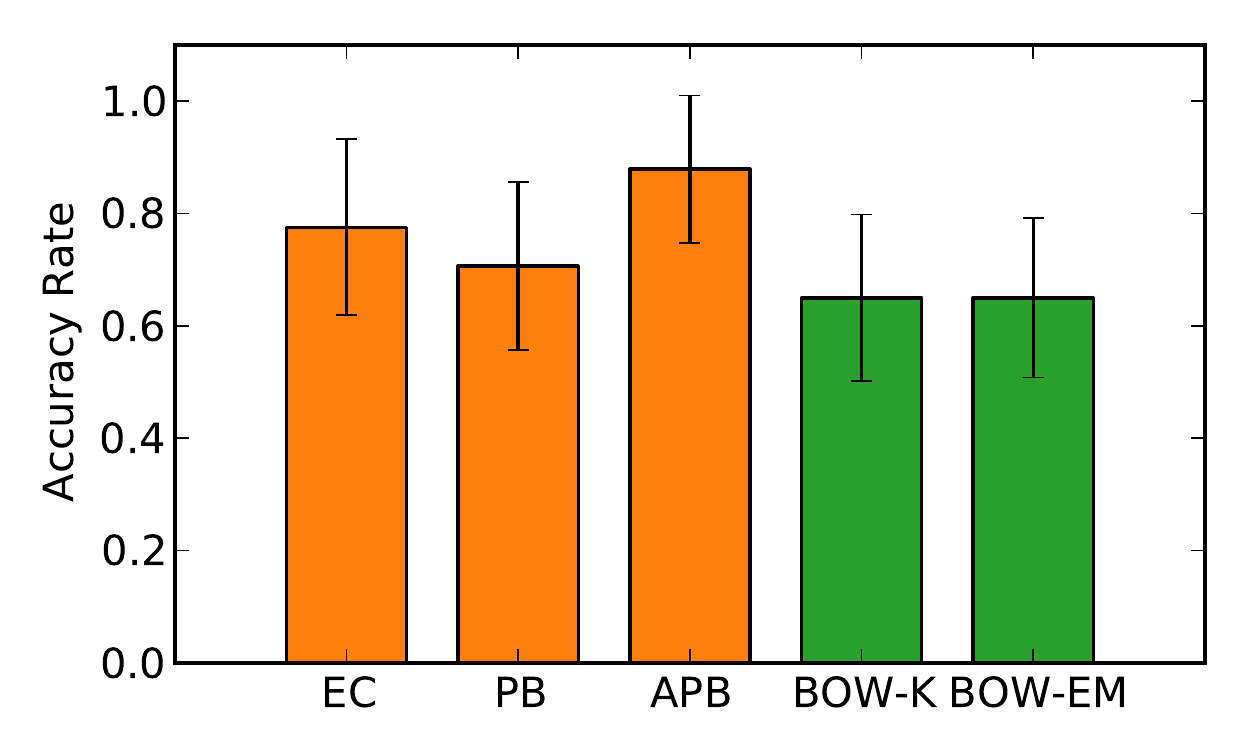}}
\subfigure[ref2][$n_s=3$ and $n_p=4$]{\includegraphics[width=5.9cm]{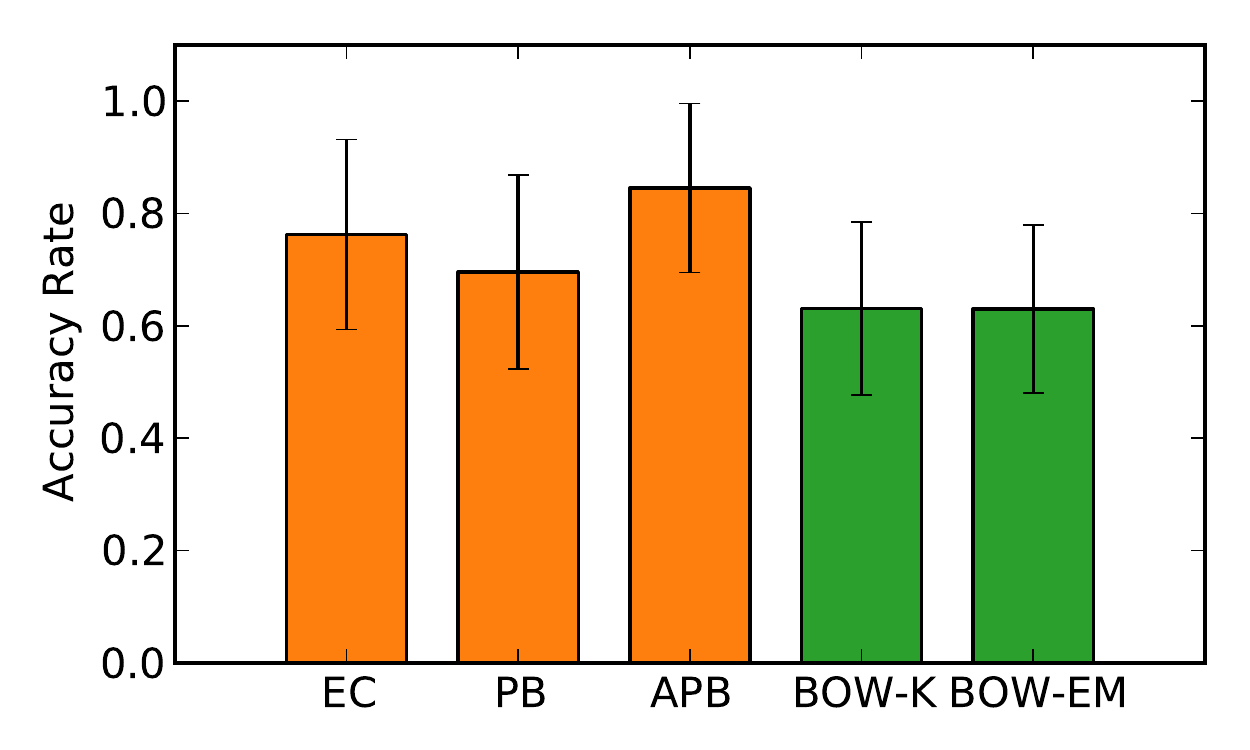}}
\subfigure[ref2][$n_s=3$ and $n_p=5$]{\includegraphics[width=5.9cm]{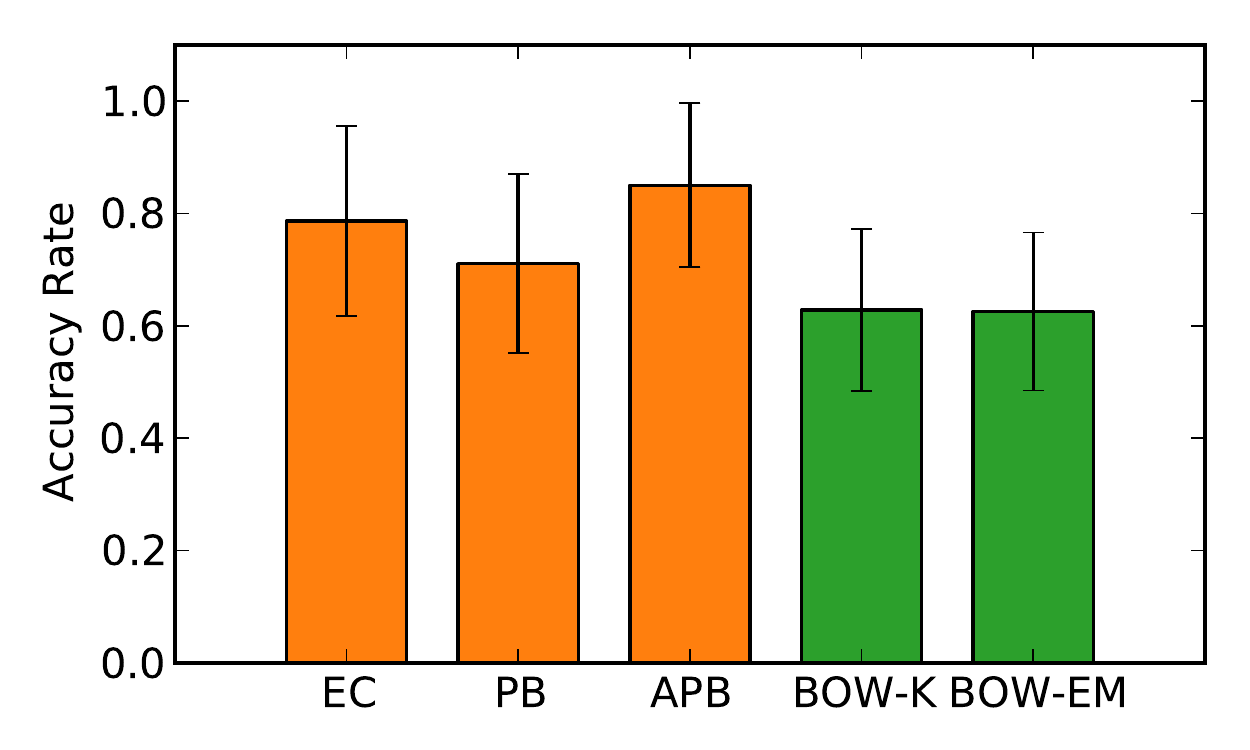}}
\subfigure[ref1][$n_s=4$ and $n_p=3$]{\includegraphics[width=5.9cm]{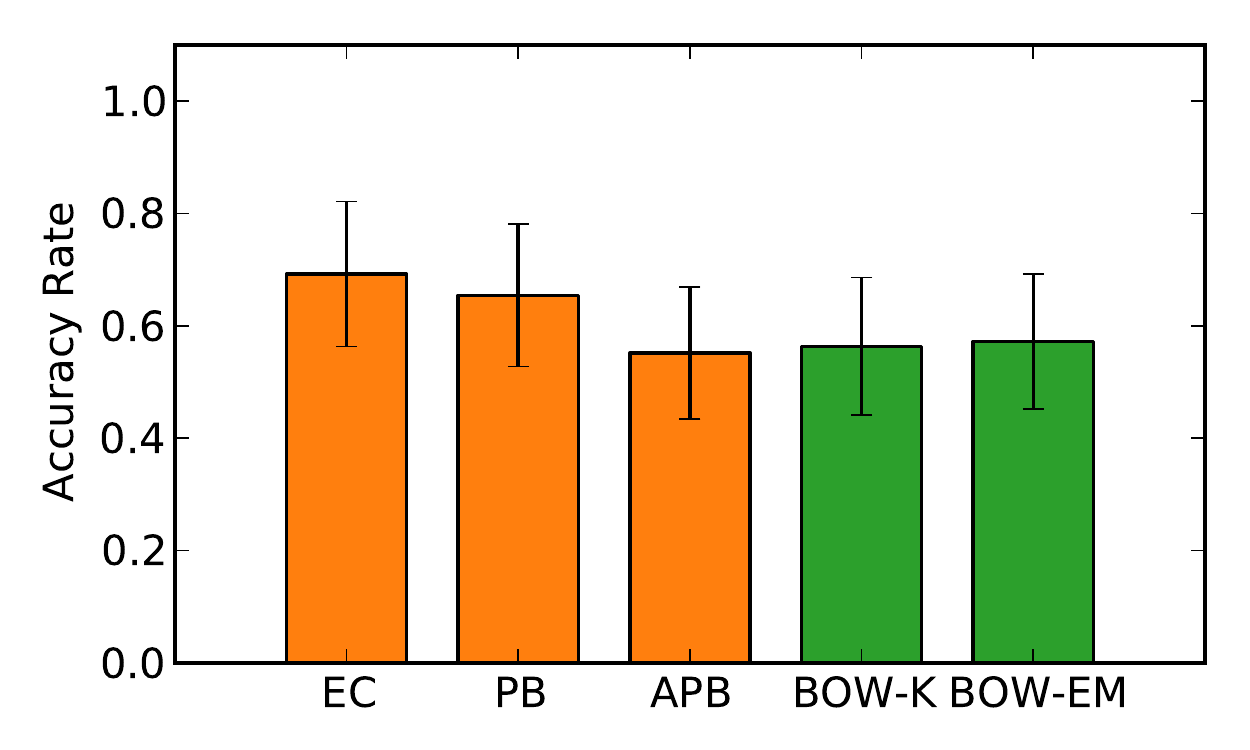}}
\subfigure[ref2][$n_s=4$ and $n_p=4$]{\includegraphics[width=5.9cm]{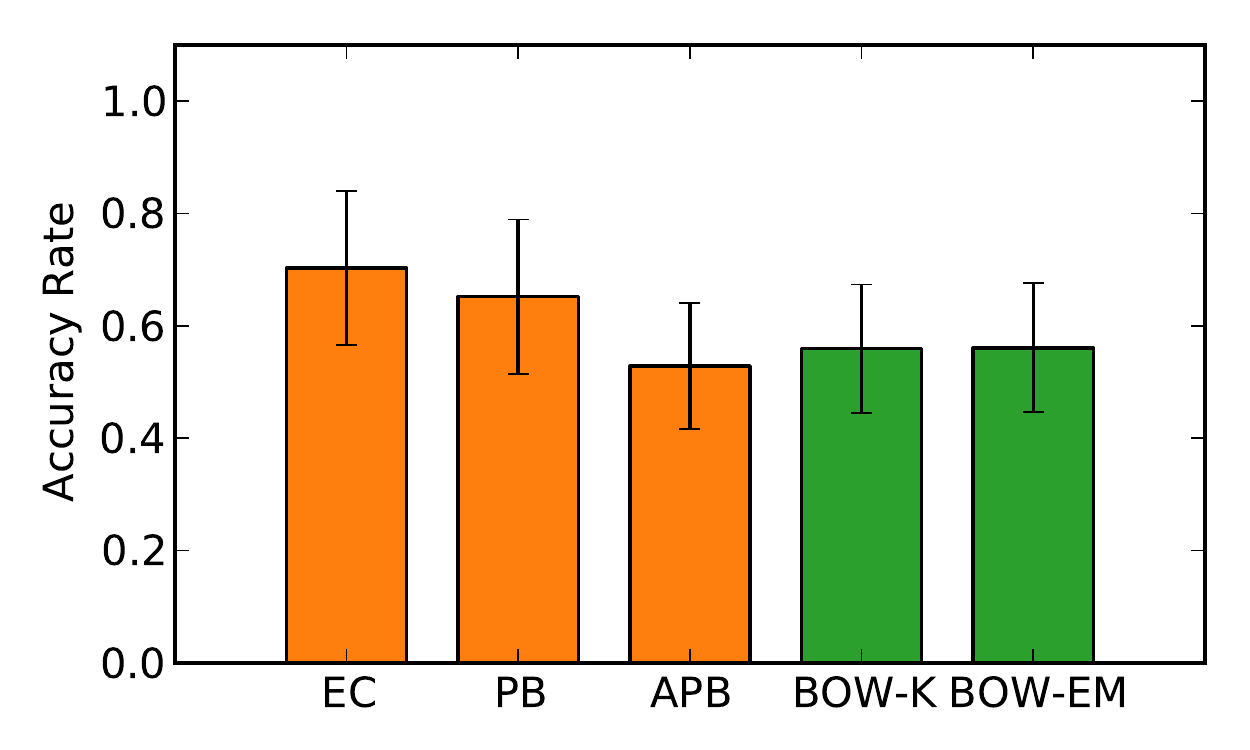}}
\subfigure[ref2][$n_s=4$ and $n_p=5$]{\includegraphics[width=5.9cm]{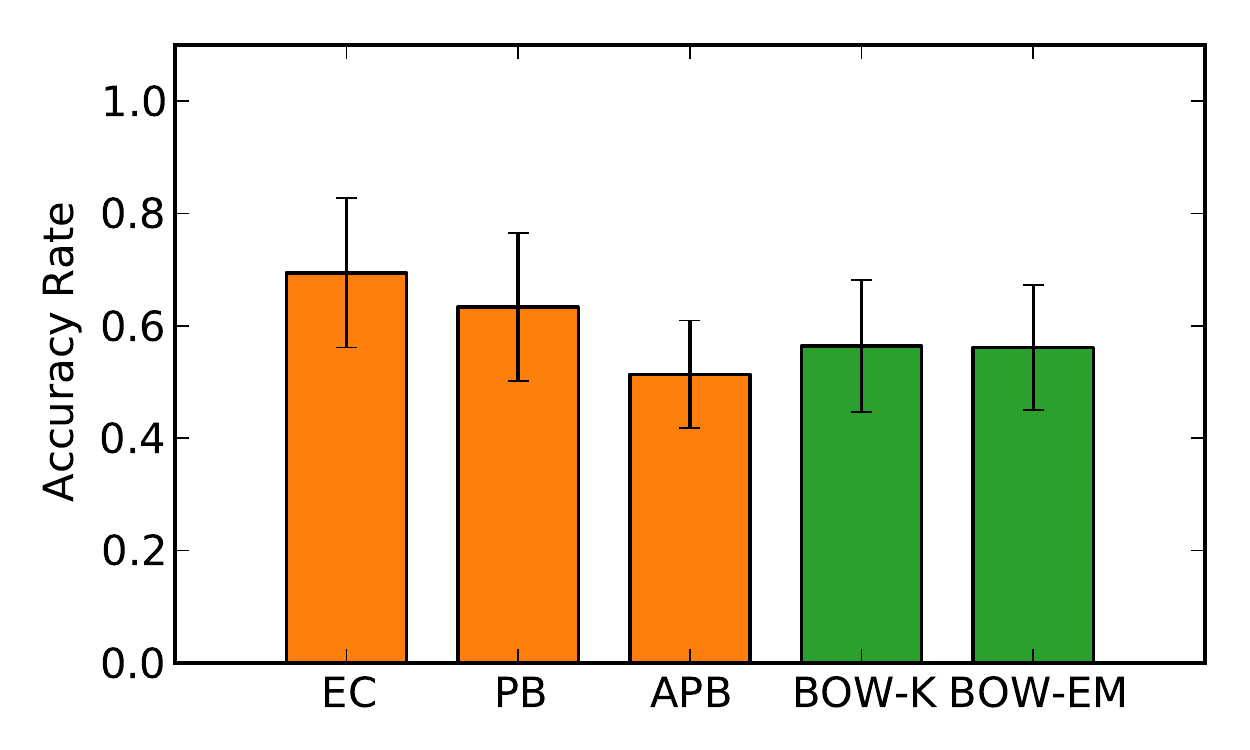}}
\caption{Performance in segmenting subjects in texts with subtopics on the same major topic. The parameters employed to generate the dataset ($n_s$, the number
of subtopics and $n_P$, the number of paragraphs per subtopic) are shown in the figure. Note that, in most of the cases, the best performance is obtained with
the APBM approach.}
\label{fig:mod2}
\end{figure*}

In Figure \ref{fig:mod3}, we show the performance obtained in the dataset comprising texts with subtopics belonging to the same major topic ({FDS}
dataset). When one compares the  networked approaches with BOW-K and BOW-EM, we note again that at least one networked approach outperformed  both
non-networked strategies. This result confirms that the networked method seems to be useful especially when the subtopics are not very different from each
other, which corresponds to the scenario found in most real documents.  The relevance of our proposed methods is specially clear when $n_s=3$.  A systematic
comparison of all three networked methods reveals that the average performance depends on specific parameters of the dataset. In most of the cases, however, the
best average performance was obtained with the EC method. The APB method also displayed high levels of accuracy, with no significant difference of performance
in comparison with EC in most of the studied cases.

\begin{figure*}
\center
\subfigure[ref1][$n_s=2$ and $n_p=3$]{\includegraphics[width=5.9cm]{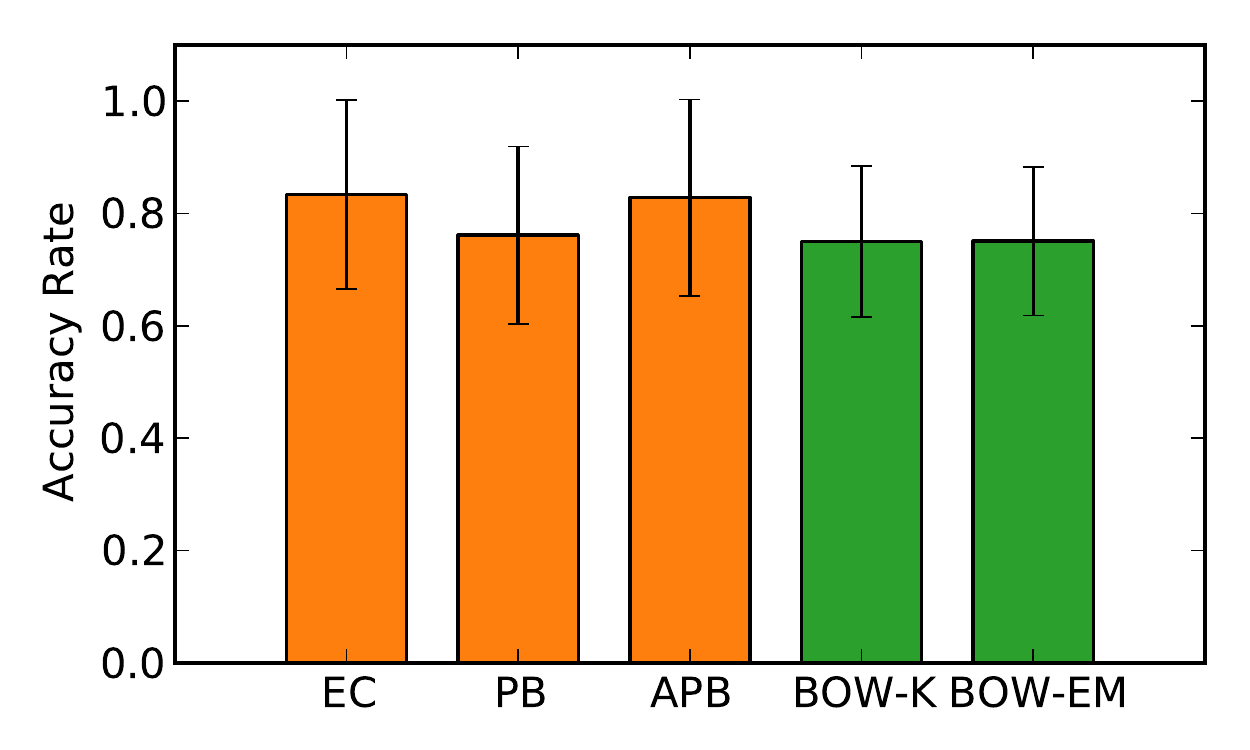}}
\subfigure[ref2][$n_s=2$ and $n_p=4$]{\includegraphics[width=5.9cm]{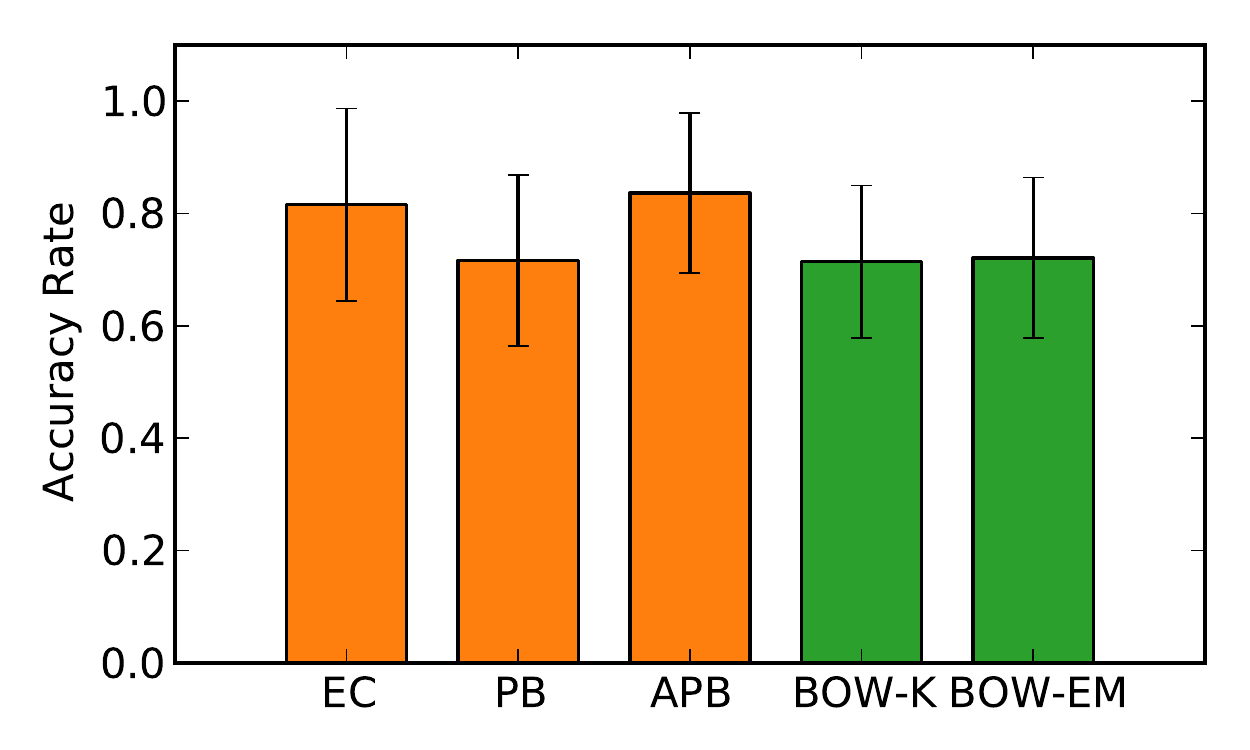}}
\subfigure[ref2][$n_s=2$ and $n_p=5$]{\includegraphics[width=5.9cm]{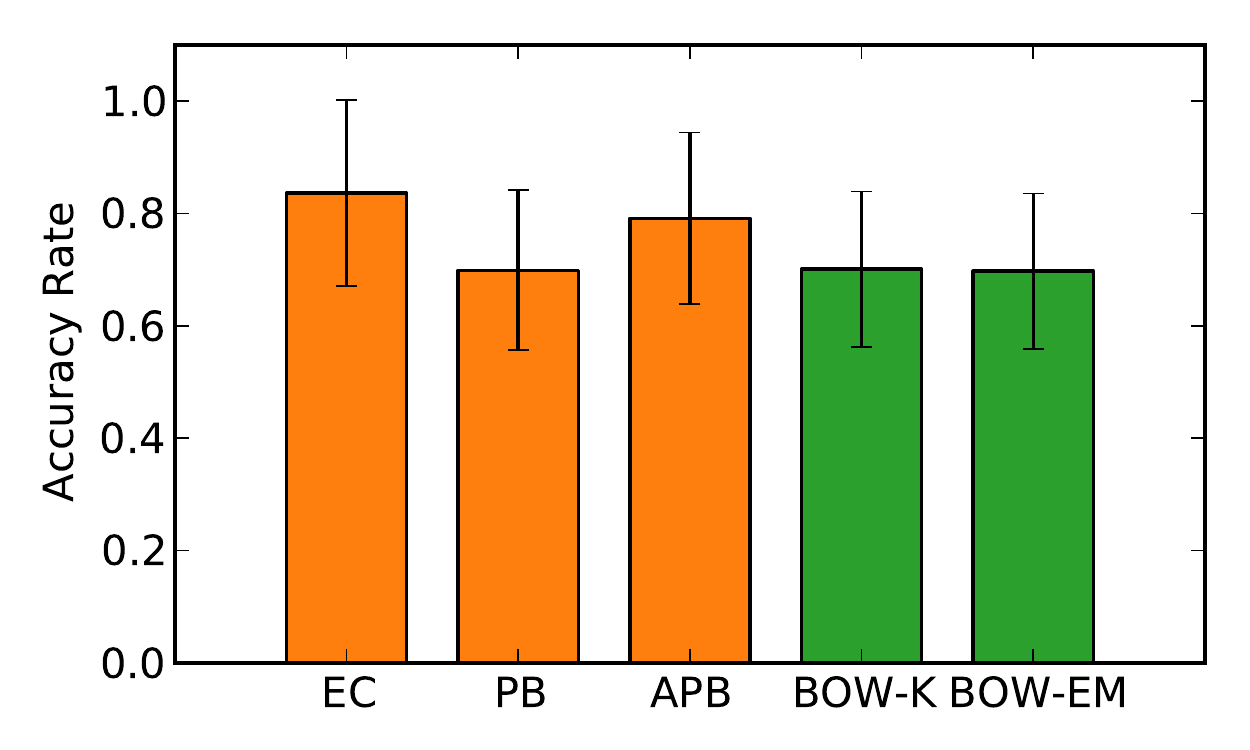}}
\subfigure[ref1][$n_s=3$ and $n_p=3$]{\includegraphics[width=5.9cm]{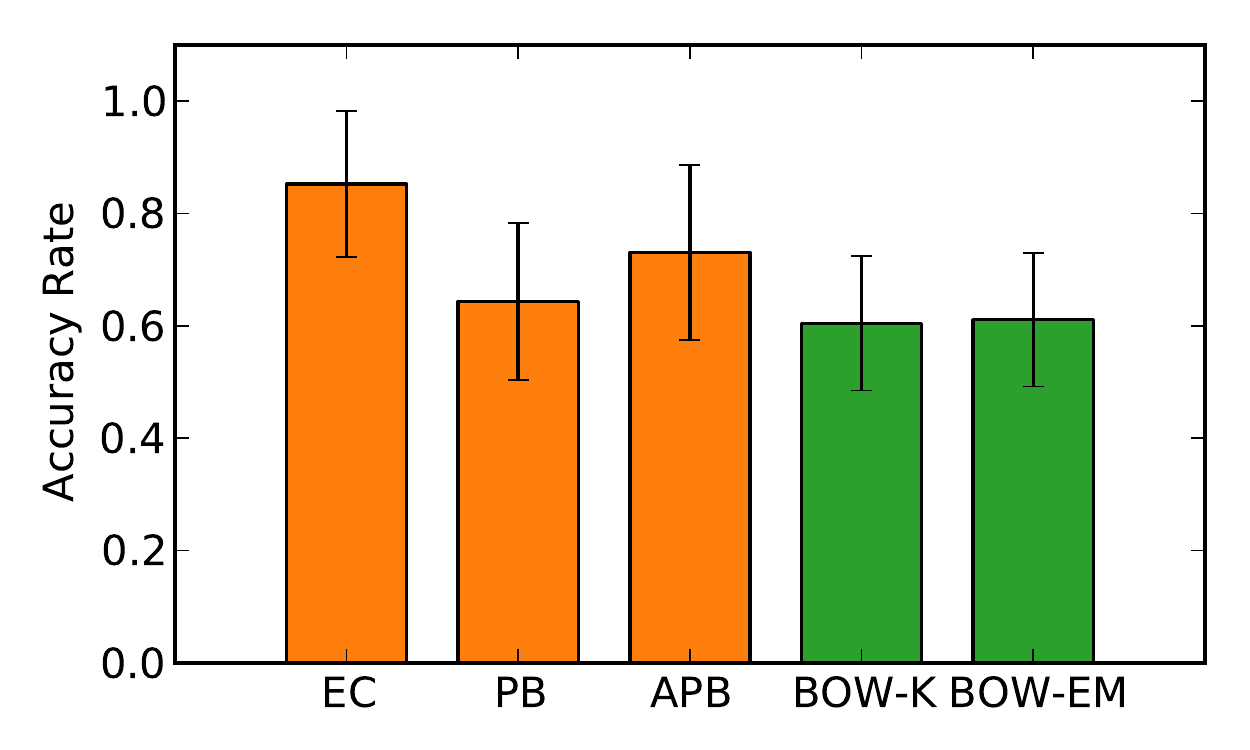}}
\subfigure[ref2][$n_s=3$ and $n_p=4$]{\includegraphics[width=5.9cm]{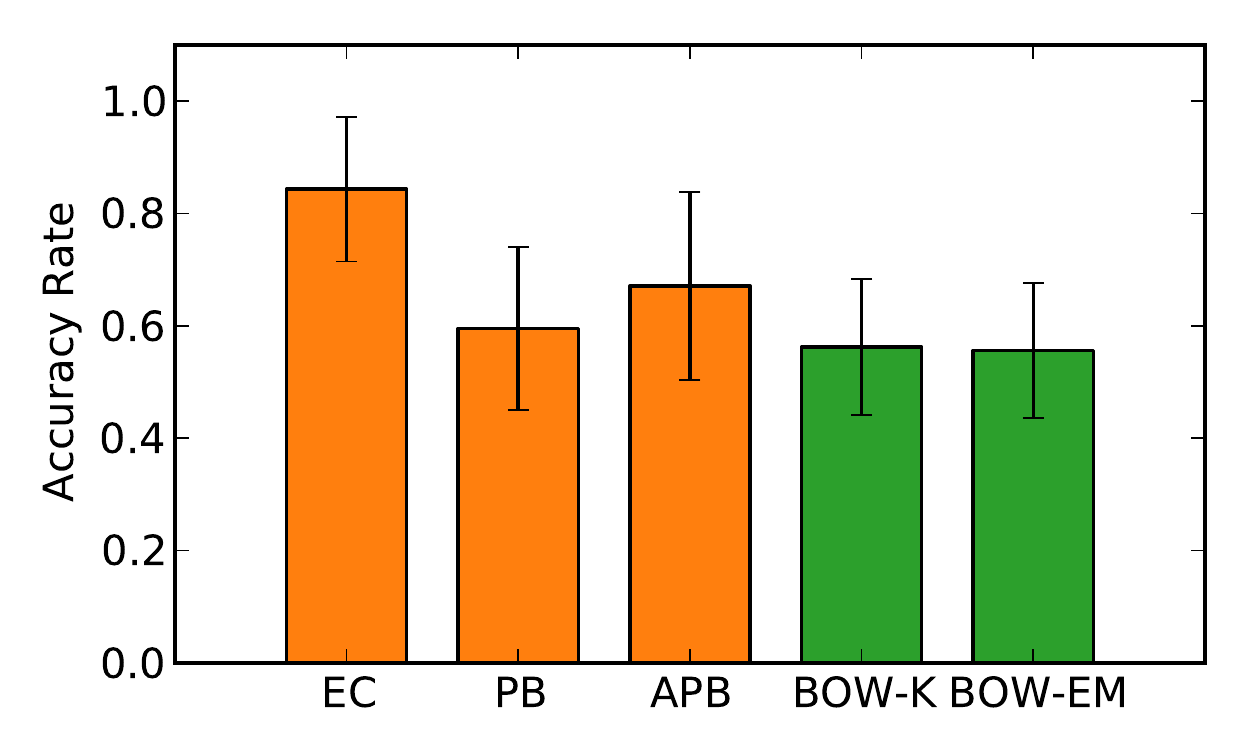}}
\subfigure[ref2][$n_s=3$ and $n_p=5$]{\includegraphics[width=5.9cm]{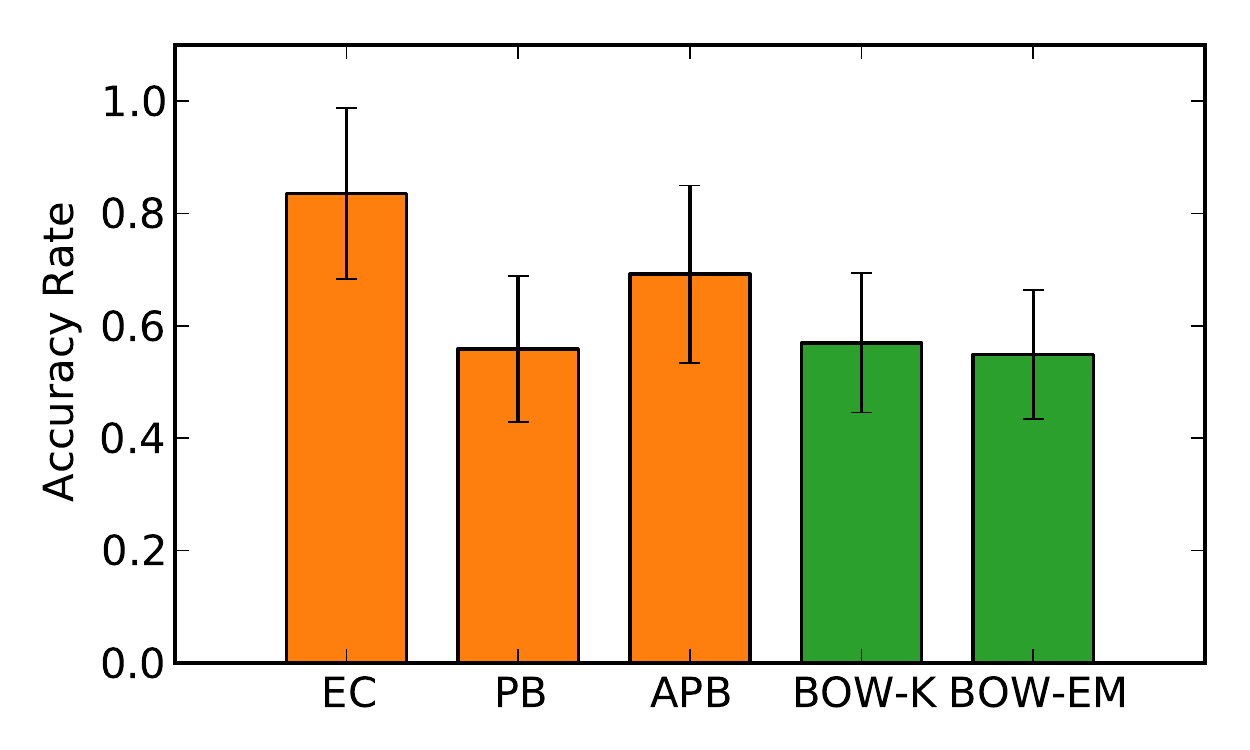}}
\subfigure[ref1][$n_s=4$ and $n_p=3$]{\includegraphics[width=5.9cm]{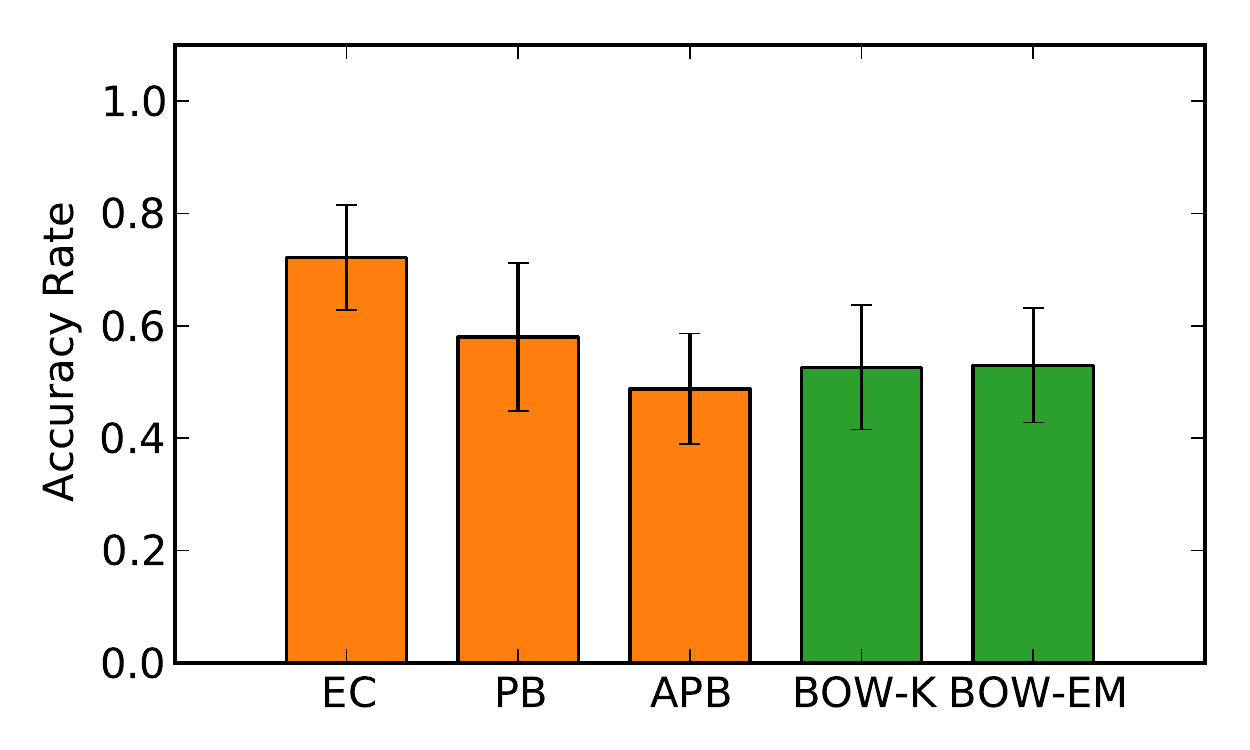}}
\subfigure[ref2][$n_s=4$ and $n_p=4$]{\includegraphics[width=5.9cm]{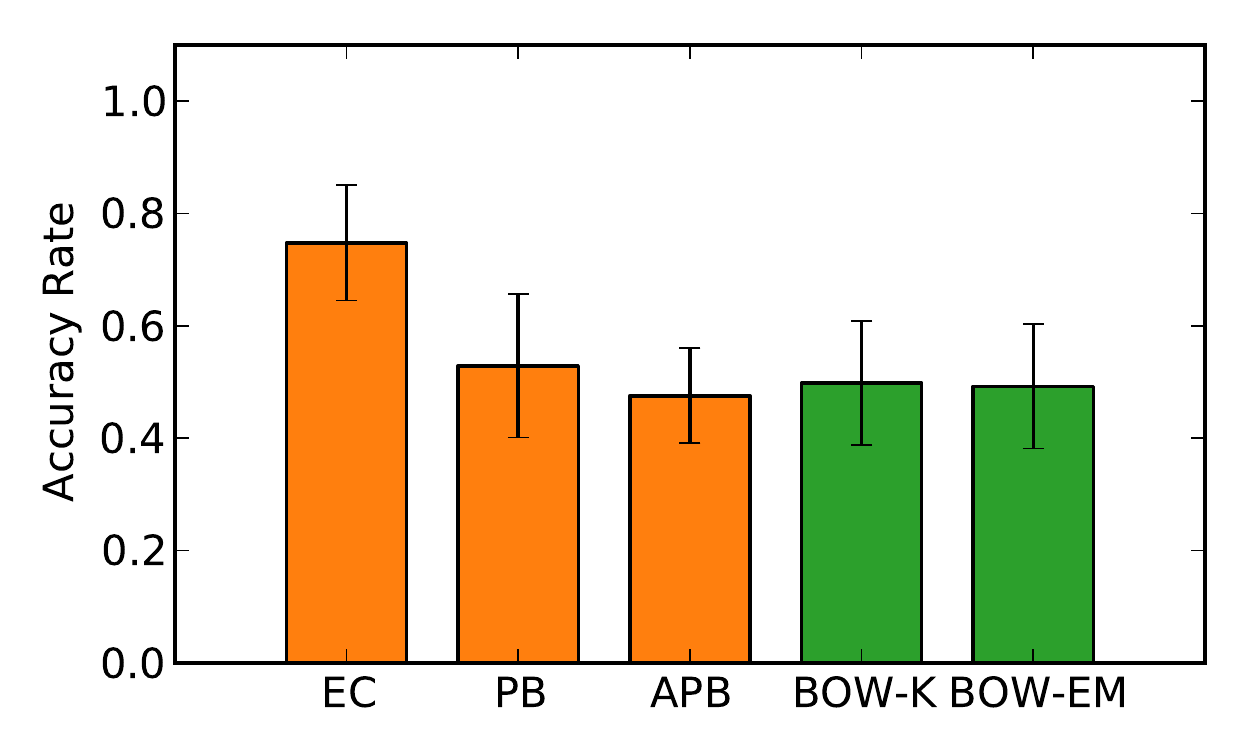}}
\subfigure[ref2][$n_s=4$ and $n_p=5$]{\includegraphics[width=5.9cm]{tds4x5.pdf}}
\caption{Performance obtained in the {FDS} dataset. The parameters employed to generate the dataset ($n_s$, the number of subtopics and $n_P$, the
number of paragraphs per subtopic)  are shown in the figure. In most of the cases, the best performance was obtained with the {EC approach}.}
\label{fig:mod3}
\end{figure*}

All in all, we have shown that networked approaches performs better than traditional techniques that do not rely on a networked representation of texts.
Interestingly, a larger gain in performance was obtained in the TSS dataset, where the difference in subtopics is much more subtle. This result suggests that
the proposed networked approaches are more suitable to analyze real texts, as one expects that the changes in subjects in a given text is much more
subtle than variations of subjects across distinct texts. Another interesting finding concerns the variations of performance in distinct datasets. Our results
revealed that there is no unique networked approach able to outperforms strategies in all studied cases. However, we have observed that, in general, both
EC and APB methods performs better than the AP approach and, for this reason, they should be tried in real applications.

\section{Conclusion}
\label{sec:conc}

In this paper, we have proposed a method to find subtopics in written texts using the structure of communities obtained in word networks.
Even though texts are written in a modular and hierarchical manner~\cite{Alvarez-Lacalle23052006}, such a feature is hidden in traditional networked text
models, as it is the case of syntactical and word adjacency networks. In order to capture the topic structure of a text, we devised three methods to link words
appearing in the same semantical context, whose length is established via parametrization. In preliminary experiments, we have found that the modular
organization is optimized when one considers a context comprising  20 words. Above this threshold, the gain in modularity was found to be not significant. We
applied our methods in a subset of articles from Wikipedia with two granularity levels.  Interestingly, in all considered datasets, at least one of the proposed
networked methods outperformed traditional bag-of-word methods. As a proof of principle, we have shown that the information of network connectivity hidden in
texts might be used to unveil their semantical organization, which in turn might be useful to improve the applications relying on the accurate characterization
of textual semantics.

In future works, we intend to use the proposed characterization to study related problems in text analysis. The high-level representation could be employed in a
 straightforwardly manner to recognize styles in texts, as one expects that distinct writing styles may approach different subjects in a very particular way. As
a consequence, the proposed networked representations could be used for identifying authors in disputed documents.  We also intend to extend our methods to make
it suitable to analyze very large documents. In this context, an important issue concerns the accurate choice  of values for the context length, which plays an
important role in the performance.   Our methods could also be combined with other traditional networked representation to improve the characterization of
related systems. Thus, a general framework could be created to study the properties of written texts in a multi-level way, in order to capture the topological
properties of networks formed of simple (e.g. words) and more complex structures (e.g. communities).


\section{acknowledgments}

Henrique Ferraz de Arruda thanks Federal Agency for Support and Evaluation of Graduate Education (CAPES-Brazil) for financial support. Diego Raphael Amancio
acknowledges S\~ao Paulo Research Foundation (FAPESP) (grant no. 2014/20830-0) for financial support. Luciano da Fontoura Costa is grateful to CNPq (grant no.
307333/2013-2), FAPESP (grant no. 11/50761-2), and NAP-PRP-USP for sponsorship.

\newpage

\bibliography{manuscript}

\end{document}